\documentclass[10pt,twocolumn,letterpaper]{article}

\usepackage{iccv}
\usepackage{times}
\usepackage{epsfig}
\usepackage{graphicx}
\usepackage{amsmath}
\usepackage{amssymb}
\usepackage{float}
\usepackage{booktabs}
\usepackage{makecell}

\usepackage[pagebackref=true,breaklinks=true,letterpaper=true,colorlinks,bookmarks=false]{hyperref}
\makeatletter
\newcommand{\printfnsymbol}[1]{%
  \textsuperscript{\@fnsymbol{#1}}%
}
\makeatother

\iccvfinalcopy 

\def\iccvPaperID{7482} 

\ificcvfinal\fi

\begin{document}

\title{Learning specialized activation functions with the Piecewise Linear Unit}

\author{
Yucong Zhou\thanks{equal contribution}\\
Huawei\\
{\tt\small zhouyucong1@huawei.com}
\and
Zezhou Zhu\printfnsymbol{1}\\
Beijing University of \\Posts and Telecommunications\\
{\tt\small zhuzezhou@bupt.edu.cn}
\and
Zhao Zhong\\
Huawei\\
{\tt\small zorro.zhongzhao@huawei.com}
}

\maketitle
\ificcvfinal\thispagestyle{empty}\fi

\begin{abstract}
The choice of activation functions is crucial for modern deep neural networks.
Popular hand-designed activation functions like Rectified Linear Unit(ReLU) and its variants show promising performance in various tasks and models.
Swish, the automatically discovered activation function, has been proposed and outperforms ReLU on many challenging datasets.
However, it has two main drawbacks.
First, the tree-based search space is highly discrete and restricted, which is difficult for searching.
Second, the sample-based searching method is inefficient, making it infeasible to find specialized activation functions for each dataset or neural architecture.
To tackle these drawbacks, we propose a new activation function called Piecewise Linear Unit(PWLU), which incorporates a carefully designed formulation and learning method.
It can learn specialized activation functions and achieves SOTA performance on large-scale datasets like ImageNet and COCO.
For example, on ImageNet classification dataset, PWLU improves 0.9\%/0.53\%/1.0\%/1.7\%/1.0\% top-1 accuracy over Swish for ResNet-18/ResNet-50/MobileNet-V2/MobileNet-V3/EfficientNet-B0.
PWLU is also easy to implement and efficient at inference, which can be widely applied in real-world applications.
\end{abstract}

\section{Introduction}

Activation functions are fundamental components for modern deep neural networks, which affect both the expressiveness and the optimization of models.
As a crucial design choice, the combination of Rectified Linear Units(ReLU)\cite{hahnloser2000digital,jarrett2009best,nair2010rectified} with deep neural networks shows six times faster in convergence compared to tanh nonlinearties\cite{alexnet}.
With non-saturating gradients, ReLU mitigates the vanishing gradient problem and shows robust performance across different tasks and neural architectures.

Beyond ReLU, many new activation functions have been proposed like Leaky ReLU\cite{maas2013rectifier}, PReLU\cite{he2015delving}, ELU\cite{clevert2015fast}, SELU\cite{klambauer2017self}, etc.
These human-designed activation functions are either fixed or have a few learnable parameters.
All these variants are similar in shape, and their gains are not consistent across tasks which limits their applications\cite{ramachandran2017searching}.

\begin{figure}[t]
\includegraphics[width=0.47\textwidth]{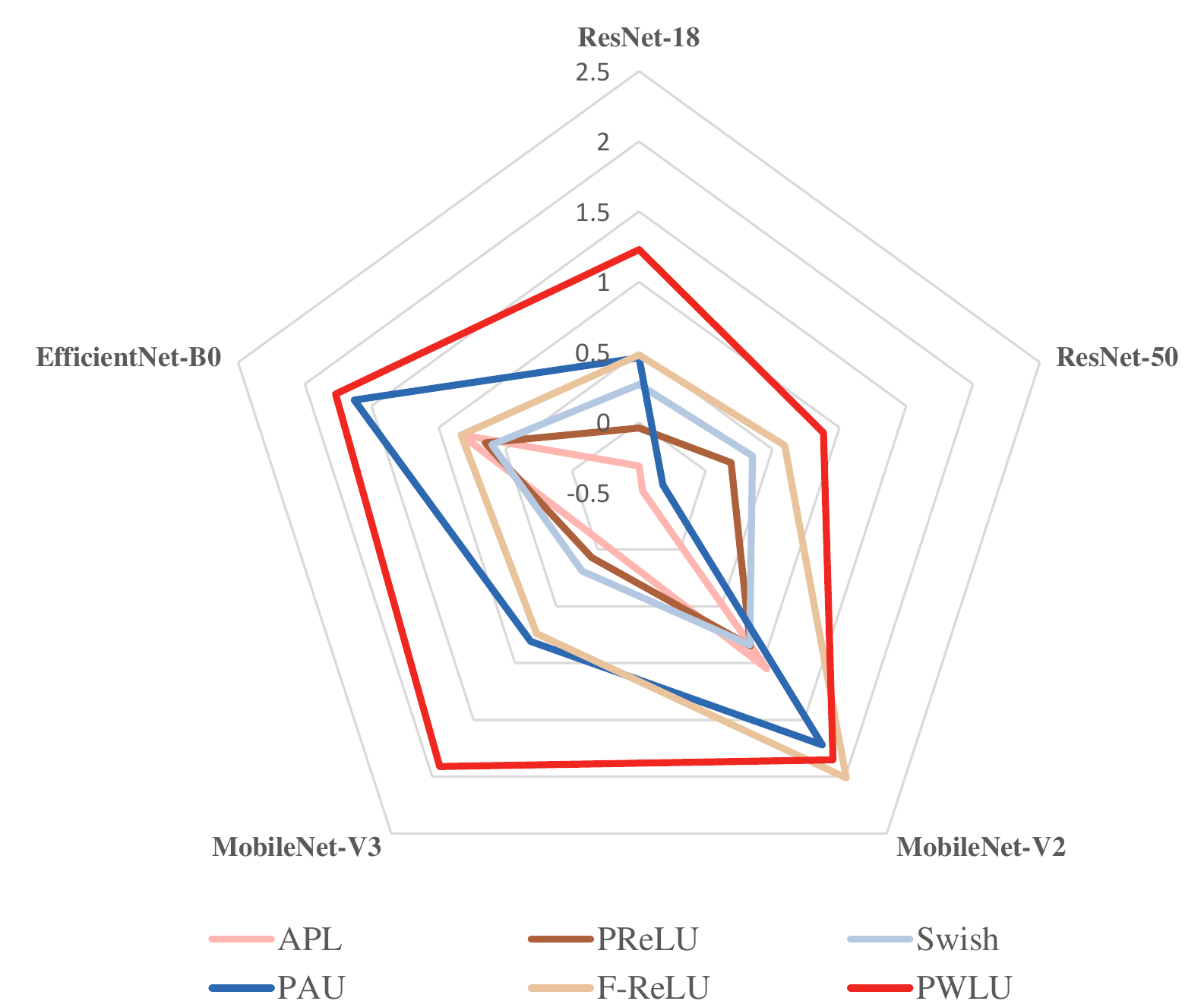}
\caption{Comparing improvements over ReLU on ImageNet dataset.
We run experiments on five different architectures for each activation function and show the improvements over ReLU baseline.
The proposed PWLU consistently outperforms other methods across architectures.}
\label{cool}
\end{figure}

Besides human-designed activation functions, automatic searched activations function, Swish\cite{ramachandran2017searching}, show better performance across tasks.
Swish adopts a tree-structured search space based on unary and binary functions, which is searched by reinforcement learning.
Although showing improvements over ReLU, Swish also has its limitations.
First, the tree-structured search space is highly discrete and restricted.
A slight change in one composing unit may result in an entirely different activation function, making it difficult to search.
Second, the sample-based searching method typically requires evaluating hundreds to thousands of candidate activation functions, which is computationally expensive.
It is infeasible to use such an inefficient searching process to find specialized activation functions for each dataset or architecture.
Instead, the searched Swish is used for most conditions.
There are also methods like APL\cite{agostinelli2014learning} and PAU\cite{molina2020pade} that use universal approximators as activation functions.
Contrary to Swish, they are mostly contiguous with respect to their parameters and can be directly optimized by gradients.
However, these methods suffer from complicated formulations that may lead to unstable learning or inefficient inference.
Besides, simply using gradients may not be sufficient to fully optimize these formulations.

To overcome these drawbacks and exploit the full potential of activation functions, we propose a new method called Piecewise Linear Unit(PWLU).
Our method is composed of two parts: the piecewise-linear-based formulation and the gradient-based learning method.
The carefully designed formulation is flexible, easy for learning and efficient for inference.
First, it covers a wide range of scalar functions which raise the potential of finding good activation functions.
Second, it is mostly differentiable with respect to its parameters and can be easily learned by gradients.
Third, its computation is very simple at inference, which is efficient and practical for real-world applications.
To effectively learn activations under this formulation, we identify the \emph{input-boundary misalignment} problem and compensate normal gradient-based learning with statistic-based realignment.
Our learning method is both effective and efficient, making it possible to learn specialized activation functions for each dataset or architecture.
With the formulation and learning method, PWLU clearly outperforms Swish and other activation functions on large-scale datasets like ImageNet\cite{russakovsky2015imagenet} and COCO\cite{lin2014microsoft}, across a wide variety of architectures.
An overall comparison on ImageNet dataset for different activation functions is shown in Figure \ref{cool}.
Particularly our method outperforms Swish for 0.9\%/0.53\%/1.0\%/1.7\%/1.0\% top-1 accuracy for ResNet-18/ResNet-50/MobileNet-V2/MobileNet-V3/EfficientNet-B0.
The learned activation functions show different preferences across architectures and layers, which confirms the benefits of learning specialized activation functions.
To summarise, our contributions are as follows:
\begin{itemize}
\item We propose a new formulation of activations based on piecewise linear functions, which is flexible, easy for learning and efficient for inference.
\item We identify the \emph{input-boundary misalignment} problem in gradient-based learning and propose a new learning method that is both effective and efficient.
\item With the proposed formulation and learning method, we demonstrate the benefits of learning specialized activation functions on large-scale datasets and different architectures.
\end{itemize}


\section{Related Work}

\subsection{Fixed-shape activation functions}

As a powerful replacement for traditional activation functions like Sigmoid and Tanh, ReLU has been widely accepted for its effectiveness and simplicity.
Many variants have been proposed to address the drawbacks of ReLU, such as the \emph{dead ReLU} problem.
Some representative activation functions include Leaky ReLU, PReLU, Softplus\cite{nair2010rectified}, ELU, SELU, etc.
While some of them show improvements in certain cases, the improvements tend to be inconsistent across different datasets and architectures.
As one possible reason, all these hand-designed activation functions are fixed in shape or parameterized by only a single parameter.
The lack of flexibility makes these activation functions only suitable for limited scenarios.

\subsection{Flexible activation functions}
To find novel activation functions, \cite{ramachandran2017searching} proposes to search for activation functions using automated search techniques.
They design a tree-structured search space that constructs activation functions using unary and binary functions.
An RNN controller is used to generate candidate activation functions in this space and updated by reinforcement learning.
Since this sample-based searching process is time-consuming, they restrict the evaluation of each activation function only on small datasets like Cifar.
But it is still too expensive to apply this searching process to each dataset or architecture.
As a result, only the searched activation function called Swish is used, but the searching process has not been applied to different datasets or architectures up to our knowledge.
Thus Swish is actually a fixed-shape activation function in most cases.

Except for sample-based method, there are also gradient-based methods such as APL\cite{agostinelli2014learning} and PAU\cite{molina2020pade}.
These methods use formulations with many learnable parameters, covering a wide range of different activation functions.
These parameters are optimized directly by gradients, which is more efficient than sampled-based searching.
Our method shares the same spirit, but the designs are substantially different.
APL uses a weighted sum of multiple ReLU-like functions which also results in piecewise-linear.
However, the formulation of our method is entirely different, which influences the corresponding gradients, learning process, and results.
PAU uses Pad\'e approximant which can also cover a wide range of activation functions.
However, the formulation is complicated and may introduce instability to optimization.
Both APL and PAU do not consider the alignment between input distribution and the parametrized functions, which is explicitly handled in our method.
Considering the computation efficiency, Both APL and PAU require much more computation than our method, limiting their applications.

\subsection{Contextual-based activation functions}

Former discussed activation functions are all scalar(one-to-one) functions(\cite{goodfellow2013maxout,hochreiter1997long,srivastava2015highway,van2016conditional,dauphin2017language}).
Recently, some contextual-based methods have been proposed, which use many-to-one functions.
For example, Dynamic-ReLU\cite{chen2020dynamic} generates the parameters of activation function based on the global context.
That is, each output element of Dynamic-ReLU depends on all input elements.
Funnel-ReLU\cite{ma2020funnel} is in the form of $y=\text{max}(x, \mathbb{T}(x))$, where $\mathbb{T}(x)$ is a spatial condition.
Funnel-ReLU also uses contextual information, but each output element only depends on a local window of $x$.
Both Dynamic-ReLU and Funnel-ReLU are dynamic to some extent.
Dynamic-ReLU has different functions for different input samples, while the functions of Funnel-ReLU are different for every single element.
These dynamic characteristics increase models' capacity and greatly improve performance.
While we focus on scalar function methods in this paper, adding dynamic mechanisms to PWLU is an interesting future direction.


\section{Methods}

\subsection{Definition of the Piecewise Linear Unit} \label{definition}
\begin{figure}[h]
\includegraphics[width=0.48\textwidth]{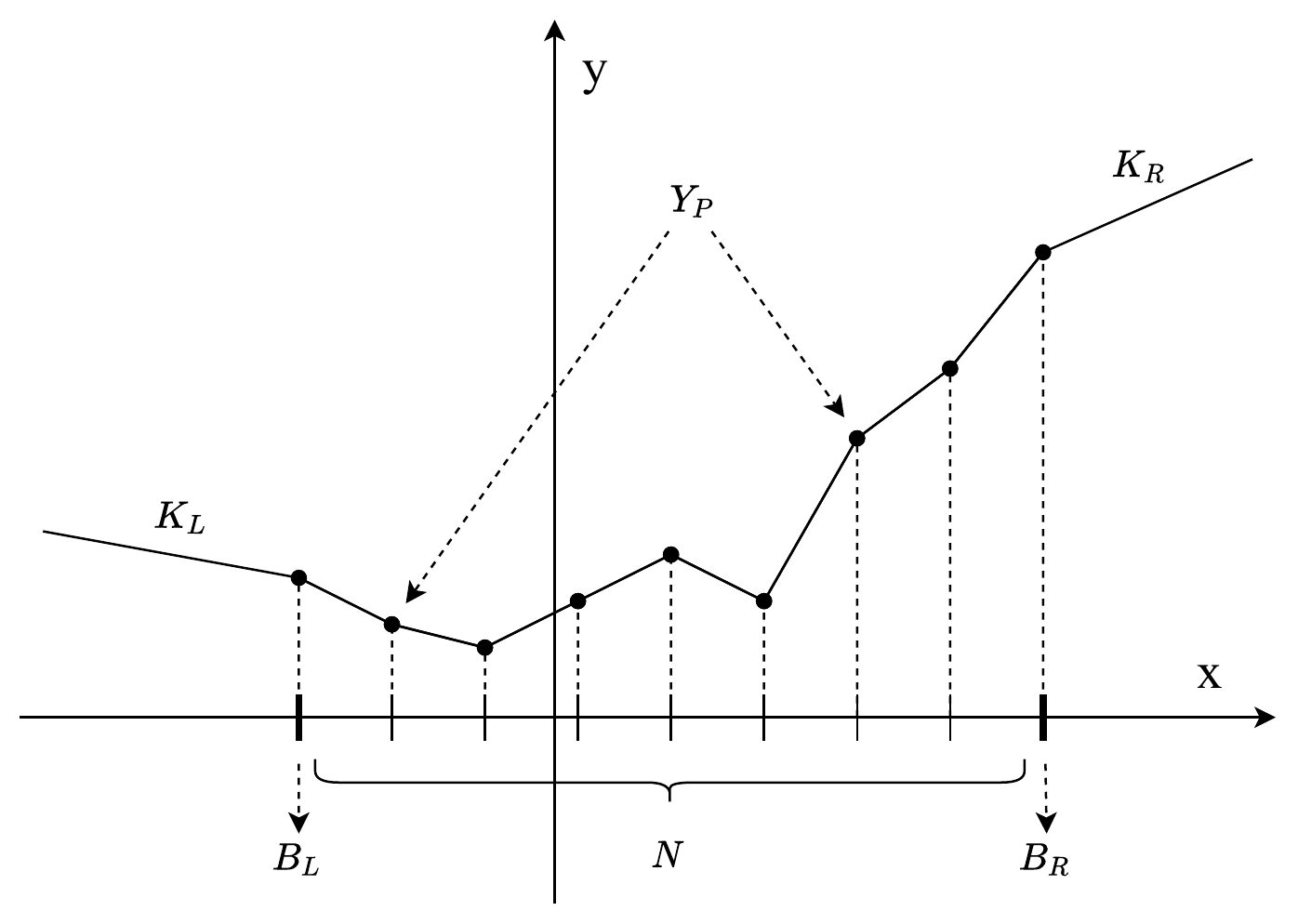}
\caption{Parameters of Piecewise Linear Unit}
\label{pwl-def}
\end{figure}

We first define the proposed activation function: Piecewise Linear Unit.
As shown in Figure \ref{pwl-def}, PWLU is defined by following parameters:
\begin{itemize}
\item Number of intervals $N$
\item Left boundary $B_L$, right boundary $B_R$
\item y-axis value of $N+1$ demarcation points $Y_P$
\item Left most slope $K_L$, right most slope $K_R$
\end{itemize}

The number of intervals $N$ is a hyperparameter, which influences the fitting ability of PWLU.
A larger number of intervals brings higher degrees of freedom, increasing the capacity of the model.
The left and right boundary $B_L, B_R$ defines the effective region where PWLU mainly focuses on.
$\left[B_L, B_R\right]$ is \emph{uniformly} divided into $N$ intervals, leaving $N+1$ demarcation points.
Each demarcation point has a corresponding y-axis value $Y_P$, which determines the shape of PWLU.
Beyond $\left[B_L, B_R\right]$, we use two slopes $K_L$ and $K_R$ to control the region out of the boundaries.

Given these parameters, the forward pass of PWLU is shown in Equation \ref{eq-definition}:
\begin{equation}
\begin{split}
\text{PWLU}_N(x, B_L, B_R, Y_P, K_L, K_R) = \\
\begin{cases}
(x-B_L)*K_L + Y_P^0 & x < B_L \\
(x-B_R)*K_R + Y_P^{N} & x \geq B_R \\
(x-B_{idx})*K_{idx} + Y_P^{idx} & B_L \leq x < B_R
\end{cases}
\end{split}
\label{eq-definition}
\end{equation}
where $idx$ indicates the index number of the interval that $x$ belongs to, while $B_{idx}$ and $K_{idx}$ are the corresponding left boundary and slope of the interval.
With $d=\frac{B_R-B_L}{N}$ denoting the interval length, these values are computed as follows:
\begin{align}
idx & = \lfloor \frac{x-B_L}{d} \rfloor \label{compute-index}\\
B_{idx}& = B_L + idx * d \\
K_{idx}& = \frac{ Y_P^{idx+1} - Y_P^{idx} }{d}
\end{align}

With this definition, PWLU has several good properties:
\begin{itemize}
\item As a universal approximator, PWLU can closely approximate any contiguous, bounded scalar functions.
\item PWLU changes contiguously with its parameters(except for hyperparameter $N$), which is friendly for gradient-based optimization.
\item Most of the flexibilities lie in a bounded region, which can maximize the utilization of learnable parameters.
\item Thanks to the uniform division of intervals, PWLU is efficient in computation, especially for inference. We will address this point in Sec \ref{speed}.
\end{itemize}

\subsection{Gradients of the Piecewise Linear Unit}

As can be seen in the previous section, the forward pass of PWLU is mostly differentiable, which means it can be directly optimized by gradients.
We derive the gradients in Table \ref{grad}.
\renewcommand{\arraystretch}{1.5}
\begin{table}[h]
\centering
\begin{tabular}{|c|c|c|c|}
\hline
    $\partial \text{PWLU}_N/ $                      & $x<B_L$     & $B_L\leq x < B_R$               & $x\geq B_R$ \\ \hline
$\partial x$           & $K_L$       & $K_{idx}$                       & $K_R$       \\ \hline
$\partial B_L$         & $-K_L$      & $K_{idx}*\frac{x-B_R}{B_R-B_L}$ & 0           \\ \hline
$\partial B_R$         & 0           & $K_{idx}*\frac{B_L-x}{B_R-B_L}$ & $-K_R$      \\ \hline
$\partial K_L$         & $x-B_L$     & 0                               & 0           \\ \hline
$\partial K_R$         & 0           & 0                               & $x-B_R$     \\ \hline
$\partial Y_P^{idx}$   & 0           & $\frac{B_{idx+1}-x}{d} $        & 0           \\ \hline
$\partial Y_P^{idx+1}$ & 0           & $\frac{x-B_{idx}}{d} $          & 0           \\ \hline
$\partial Y_P^{0}$     & 1           & 0                               & 0           \\ \hline
$\partial Y_P^{N}$     & 0           & 0                               & 1           \\ \hline
\end{tabular}
\vspace{0.2cm}
\caption{Gradients of Piecewise Linear Unit}
\label{grad}
\end{table}

Note that in the derivation we treat $idx$ as constant and omit related gradients.
For gradients with respect to $Y_P$, we separate it into four parts for clarity.
When $x \in \left[B_{idx}, B_{idx+1}\right]$ where $0 \leq idx \leq N$, the output of PWLU has gradient with respect to both $Y_P^{idx}$ and $Y_P^{idx+1}$.
When $x \in \left[-\infty, B_L\right]$, the output of PWLU only has gradient with respect to $Y_P^0$.
When $x \in \left[B_R, \infty \right]$, the output of PWLU only has gradient with respect to $Y_P^N$.
As can be seen in the definition and gradients, the computation of PWLU is straightforward.
We implement the forward and backward computation in CUDA for efficiency.

\subsection{Learning the Piecewise Linear Unit}
\label{sec_search}

Given the definition and the gradients of PWLU, we can learn the parameters of PWLU by gradient descent directly.
Before learning, the parameters of the PWLU should be properly initialized to stabilize training.
One straightforward way is to initialize PWLU to existing activation functions such as ReLU.
For example, if $N$ is an even number, we can initialize $B_L$ to any negative value and set $B_R=-B_L$, $K_L=0$, $K_R=1$, then set each $Y_P^{idx}=\text{ReLU}(B_{idx})$.
This initialization offers a good start, but there is a further problem that needs to be handled during training.

\vspace{-0.3cm}

\begin{figure}[h]
\centering
\includegraphics[width=0.45\textwidth]{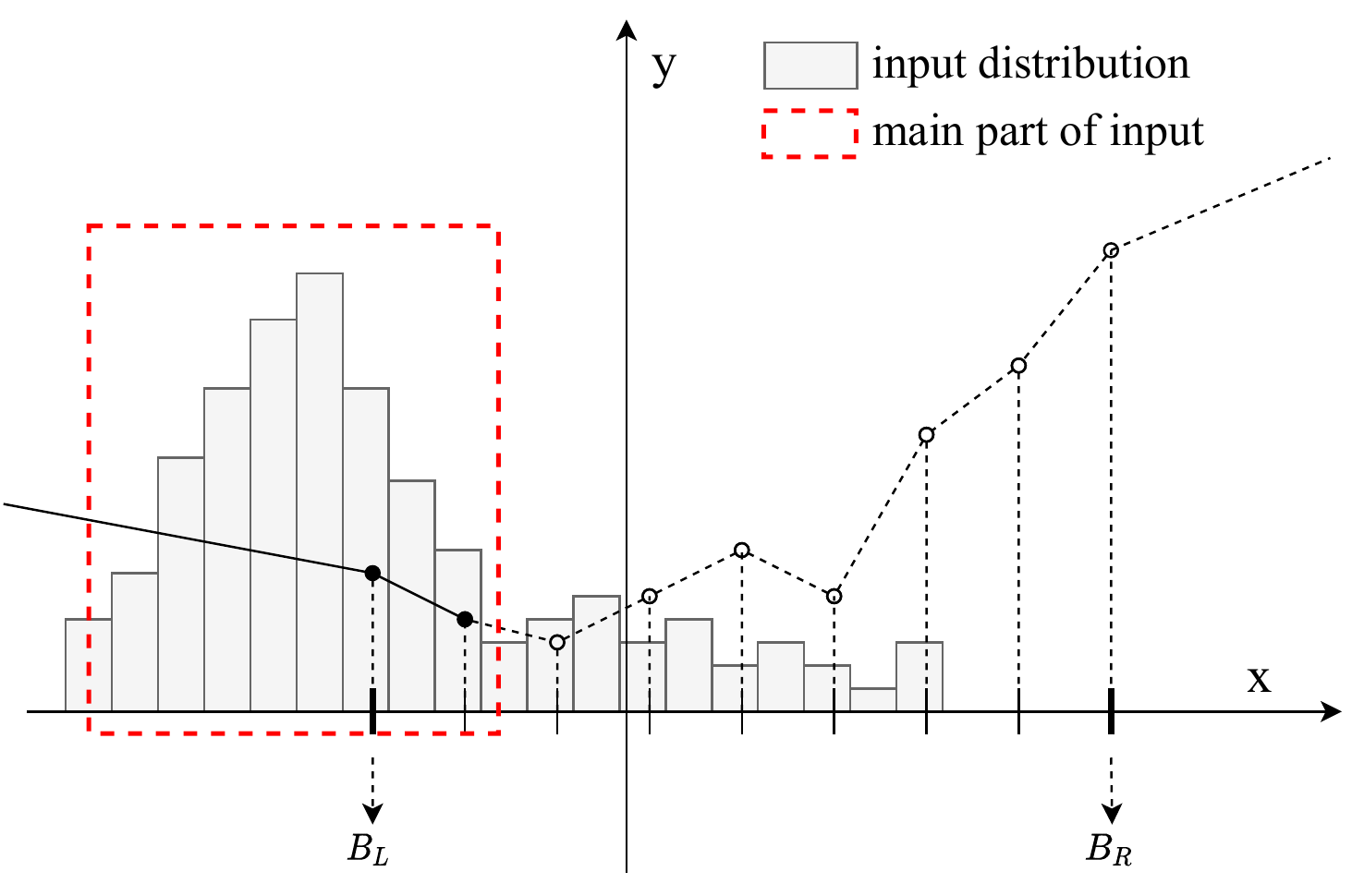}
\caption{Illustration of the misalignment between $\left[B_L, B_R\right]$ and the input distribution}
\label{mismatch-problem}
\end{figure}

\paragraph{Input-boundary misalignment} $B_L, B_R$ are important parameters of PWLU which define the main region where the shape can be learned.
According to the definition of PWLU, the degree of freedom mainly lies in the region $\left[B_L, B_R\right]$.
There are $N+1$ parameters($Y_P$) controlling the shape of PWLU in this region, while only two parameters($K_L$, $K_R$) for $\left[-\infty, B_L\right]$ and $\left[B_R, \infty\right]$.
Intuitively this main region should be aligned with the input distribution of PWLU, otherwise the effective degrees of freedom can reduce.
Consider the situation illustrated in Figure \ref{mismatch-problem}.
The main part of the input distribution shifts to the left, which only has a little intersection with $\left[B_L, B_R\right]$.
Under this condition, the parameters out of the input distribution have little contribution to the network, wasting the flexibility of PWLU and affecting the final performance.
One may expect that gradients can drive $B_L$, $B_R$ to their optimal values, but unfortunately we find that gradients do not help either.
Gradients with respect to $B_L$ and $B_R$ only follow the direction that \emph{lowers the task loss}, which is not directly correlated with the alignment between $\left[B_L, B_R\right]$ and the input distribution.
During training, this \emph{input-boundary misalignment} problem will continuously exist and hinder the learning of PWLU.

\vspace{-0.4cm}

\begin{figure}[h]
\includegraphics[width=0.5\textwidth]{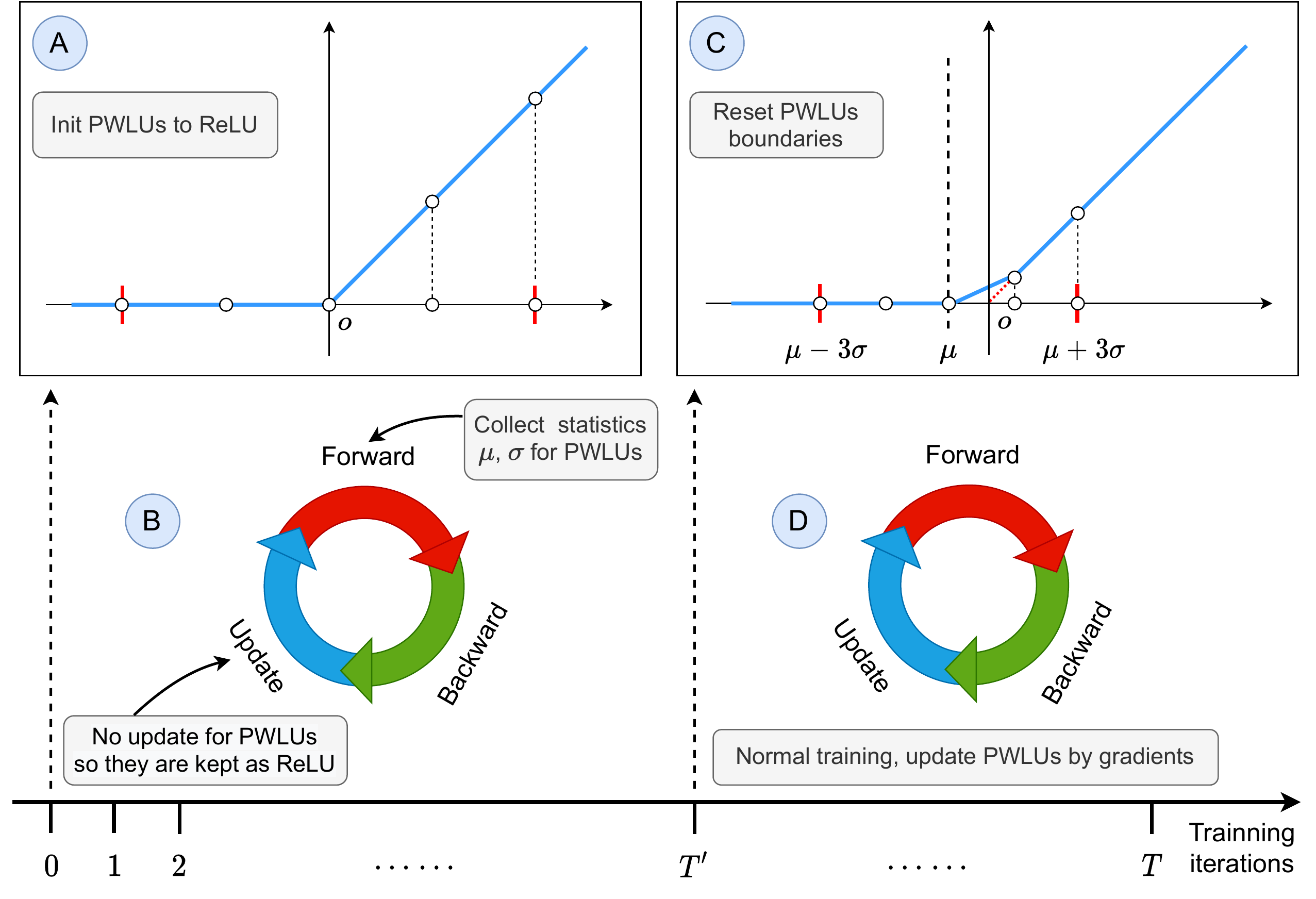}
\caption{Learning PWLU with statistic-based realignment}
\label{two-stage}
\end{figure}

\paragraph{Statistic-based realignment} To overcome this problem, we propose a statistic-based method to facilitate the learning of PWLU.
The training process is split into two phases and the full pipeline is shown in Figure \ref{two-stage}.

\vspace{-0.4cm}

\paragraph{Phase I} Phase I lasts from iteration 0 to $T^{'}-1$. At the beginning of phase I, we first initialize all PWLUs to ReLU(Figure \ref{two-stage} A) and then start the training.
In addition to standard training, we have two modifications(Figure \ref{two-stage} B):
\begin{itemize}
\item during each forward procedure, we collect statistics of input distribution for each PWLU
\item during each update procedure, we do not update the parameters of PWLUs to keep them as initialized ReLU
\end{itemize}
For collecting statistics we compute the running mean and running standard deviation of input $x$ for each PWLU, as formulated in Equation \ref{eq-stat}.
These statistics will be used in Phase II.
\begin{equation}
\begin{aligned}
\mu &= \mu * 0.9 + \text{mean}(x) * 0.1 \\
\sigma &= \sigma * 0.9 + \text{std}(x) * 0.1
\end{aligned}
\label{eq-stat}
\end{equation}

\paragraph{Phase II} Phase II lasts from iteration $T^{'}$ to $T$.
At the beginning of Phase II, we first reset parameters of each PWLU as following(Figure \ref{two-stage} C):
\begin{equation}
\begin{gathered}
B_L = \mu - 3 * \sigma,\ \ \ B_R = \mu + 3 * \sigma \\
K_L = 0,\ \ \ K_R = 1 \\
Y_P^{idx} = \text{ReLU}(B_{idx}) \ \ idx \in \{0, 1, 2, ..., N\}
\end{gathered}
\label{eq-reset}
\end{equation}
Here we adopt the 3-sigma rule, which works well in our experiments.
After resetting, the PWLUs are still in ReLU form as in Phase I (there may be a tiny gap around zero point whose influence is negligible), but their boundaries are aligned with the input distribution.
Then we start normal training and update parameters of PWLUs by gradients.

\vspace{-0.3cm}

\paragraph{Summary} The proposed learning method has many advantages.
First, it is highly efficient. With gradient-based learning, it is feasible to learn specialized activation functions for each dataset, architecture, layer or even channel, at the cost of a single training.
Second, it is effective and robust. With special handling of the input-boundary misalignment problem, our method fully exploits the potential of PWLU.
Third, it is simple to implement. Our method only requires slight changes to normal training procedures, which are limited to PWLU itself and do not affect any other components.
Our method offers great compatibility with all kinds of techniques.

%
%


\section{Experiments}

We first show overall results on ImageNet and COCO in section \ref{imagenet} and section \ref{coco}, then ablation studies on several aspects of PWLU in later sections.

\subsection{Experiments on ImageNet}
\label{imagenet}

We first evaluate our method on the ImageNet classification dataset, which contains 1.28 million training images and 50k validation images.

\vspace{-0.2cm}

\paragraph{General setups}
All the architectures are trained on 16 Nvidia V100 GPUs with a total batchsize of 1024, SGD optimizer with momentum 0.9, and the one-cycle cosine learning rate scheduler.
Weight decay is set to 0.0001 for ResNet-18 and ResNet-50\cite{he2016deep}, 0.00004 for MobileNet-V2\cite{sandler2018mobilenetv2} and MobileNet-V3\cite{howard2019searching}, 0.00001 for EfficientNet-B0\cite{tan2019efficientnet}.
We use the standard random crop + flip augmentation for ResNets and MobileNets, and AutoAugment for EfficientNet, following their original implementations.
We train ResNets for 100 epochs, MobileNet-V2 for 200 epochs, MobileNet-V3 and EfficientNet-B0 for 350 epochs.

\paragraph{PWLU setups}
When applying PWLU to an architecture, we replace all activation layers except those that have tiny input feature maps, such as activations in SE-block\cite{hu2018squeeze}.
Tiny input feature maps are not suitable for learning parameters of PWLU, so we let these activations unchanged as in their original implementation.
$T^{'}$ is set to 5 epochs as it works well across our experiments.
We use the channel-wise setting for PWLU. $N$ is set to 16.

\vspace{-0.3cm}
\paragraph{Other activation functions}
We choose three types of activation functions for comparison.
\begin{itemize}
\item
The first type of activation function includes \textbf{ReLU}, \textbf{PReLU} and \textbf{Swish}, which have fixed(or nearly fixed) shape.
They are simple in formulation and widely used in different tasks.
\item
The sencond type of activation function inlcludes Adaptive Piecewise Linear(\textbf{APL}) activation and Pad\'e Activation Units(\textbf{PAU}).
We choose them for they share the same spirit of using a flexible parametric formulation.
For APL, we find that the default setting of hyperparameter $S=5$ is unstable on ImageNet.
So we set $S=3$, which is the best value that can be successfully trained.
For PAU we follow the original setting where $m=5, n=4$.
\item
For the third type of activation function we choose \textbf{Funnel-ReLU}, which is a many-to-one function.
While PWLU is a scalar function, we choose Funnel-ReLU as a representative for another line of works.
\end{itemize}

\vspace{-0.4cm}
\paragraph{Architectures} To compare the general performance of different activation functions, we conduct experiments on 5 widely used architectures including ResNet-18, ResNet-50, MobileNet-V2, MobileNet-V3 and EfficientNet-B0.
They cover a wide range of computational complexity and number of parameters that can be applied to different cases.

\begin{table*}[htbp]
\centering
\begin{tabular}{lccccc}
\toprule
                  & ResNet-18& ResNet-50   & MobileNet-V2   & MobileNet-V3& EfficientNet-B0 \\
\midrule
FLOPs             & 1.8G     & 4.1G        &  300M          & 56M         &    384M         \\
\#Params          & 11.69M   & 25.5M       &  3.5M          & 2.6M        &    5.3M         \\
\midrule
ReLU              & 71.34    & 76.90       & 72.84          & 67.74       & 76.45   \\
Swish             & 71.61    & 77.25       & 73.68          & 67.93       & 77.05   \\
PReLU             & 71.30    & 77.09       & 73.69          & 67.81       & 77.10   \\
APL               & 71.03    & 76.43       & 73.89          & 67.68       & 77.28   \\
PAU               & 71.80    & 76.58       & 74.56          & \textbf{68.55}  & \textbf{78.08}   \\
F-ReLU            & \textbf{71.82}  & \textbf{77.49} & \textbf{74.85} & 68.48       & 77.28   \\
\midrule
\textbf{PWLU}     & \textbf{72.57}  & \textbf{77.78} & \textbf{74.69} & \textbf{69.65}  & \textbf{78.22}   \\
\bottomrule
\end{tabular}
\vspace{0.2cm}
\caption{Top-1 accuracies on ImageNet for different activation functions and architectures. For each architecture the top-2 results are shown in bold.}
\label{tab-overall}
\end{table*}

\begin{figure}[h]
\includegraphics[width=0.48\textwidth]{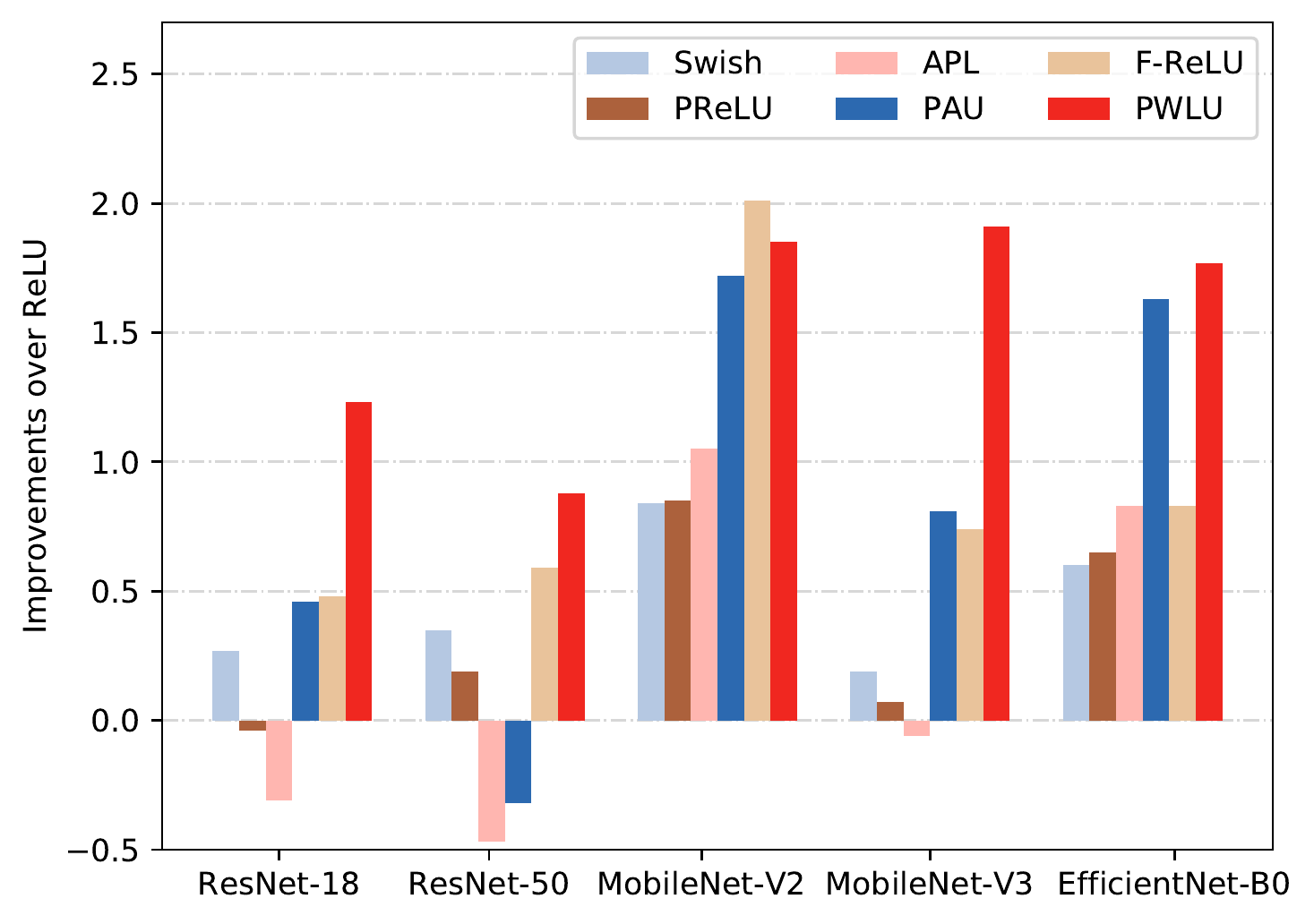}
\caption{Improvements in Top-1 accuracy over ReLU}
\label{bar}
\end{figure}

\vspace{-0.3cm}
\paragraph{Results} The top-1 accuracies are shown in Table \ref{tab-overall}.
Our method outperforms ReLU and Swish in all experiments.
To show differences more clearly, we also plot the improvements of top-1 accuracy over ReLU in Figure \ref{bar}.
Although these activation functions are better than ReLU in most cases, an interesting pattern is worth noting: the improvements are highly inconsistent across architectures, except for PWLU.
For example, Swish is claimed to perform well especially for mobile-sized models, however it is true only for MobileNet-V2 in our experiments while the improvements for ResNet-18 and MobileNet-V3 are marginal.
PReLU and APL have a similar trend with Swish but are more unstable.
PAU and Funnel-ReLU generally perform better than Swish, but still have large differences between architectures.
On the other hand, PWLU achieves the best results and performs consistently across architectures.


\subsection{Experiments on COCO}
\label{coco}
To further demonstrate the generalization ability of our method, we apply PWLU to the COCO object detection dataset.
We use the \emph{train2017} split(118k images) for training and the \emph{val2017} split(5k images) for testing.
We use two popular detection framework: Mask R-CNN\cite{he2017mask} and RetinaNet\cite{lin2017focal}.
We mainly follow the settings in \cite{he2019rethinking} which do not use ImageNet pre-train models but train the model from scratch.
The detailed settings are as follows.

\vspace{-0.3cm}
\paragraph{Architecture} We use ResNet-50 plus Feature Pyramid Network(FPN)\cite{lin2017feature} as backbone for both Mask R-CNN and RetinaNet .
SyncBN is used to facilitate the from-scratch training.
For Mask R-CNN the Region Proposal Networks(RPN)\cite{ren2015faster} is trained jointly with Mask-RCNN.

\vspace{-0.3cm}
\paragraph{Learning rate scheduling} Both Mask R-CNN and RetinaNet are trained for 270k iterations(namely, `3$\times$ schedule').
The learning rate is reduced by 10$\times$ in the last 60k and the last 20k iterations.

\vspace{-0.3cm}
\paragraph{Hyper-parameters} The initial learning rate is 0.02, weight decay is 0.0001 and momentum is 0.9.
All models are trained with 8 Nvidia V100 GPUs with two images per GPU.
The image scale is 800 pixels for the shorter side.
For PWLU we use the same setting as in ImageNet.

\vspace{-0.3cm}
\paragraph{Results} We show results for Mask R-CNN and RetinaNet with different activations in Table \ref{tab-coco}.
As Swish outperforms ReLU for 1\%/0.8\% AP, our PWLU further improves 0.4\%/0.5\% AP over Swish, which is on par with F-ReLU.
Note that F-ReLU utilizes contextual information which is likely beneficial for detection tasks, while PWLU does not access this extra information.

\begin{table}[h]
\centering
\small
\begin{tabular}{lcccc}
\toprule
                  & \multicolumn{2}{c}{Mask R-CNN}                          &  RetinaNet                \\
\midrule
                  & $\text{AP}^{\text{bbox}}$  &  $\text{AP}^{\text{mask}}$ & $\text{AP}^{\text{bbox}}$ \\
\midrule
ReLU              & 39.06  & 35.3   & 37.06  \\
Swish             & 40.05  & 36.1   & 37.82  \\
F-ReLU           & 40.25  & 36.90  & \textbf{38.35}  \\
\midrule
\textbf{PWLU}   & \textbf{40.48}  & \textbf{37.13}  & 38.31  \\
\bottomrule
\end{tabular}
\vspace{0.2cm}
\caption{Results for different activations on COCO dataset}
\label{tab-coco}
\end{table}

\subsection{Efficient inference for PWLU}
\label{speed}

As noted in section \ref{definition}, we show how to optimize efficiency for PWLU at inference.
Re-write the main computation as $x*K_{idx}+(Y_P^{idx}-B_{idx}*K_{idx})$, where $K_{idx}$ and $Y_P^{idx}-B_{idx}*K_{idx}$ can be calculated in advance.
Then the main computation becomes a single Multiply-Add: $x*S_{idx} + O_{idx}$.
The comparings in Equation \ref{eq-definition} can be eliminated as well.
The first two conditions can be merged into the third by extending $idx$.
The original $idx$ belongs to $\{0, 1, \cdots, N-1\}$, now we extend it to $\{-1, 0, \cdots, N\}$ by $idx^{'} = \text{clip}(\lfloor\frac{x-B_L}{d}\rfloor, -1, N)$.
$idx^{'}=-1$ corresponds to $x < B_L$ while $idx^{'}=N$ corresponds to $x \geq B_R$.
Then we set $S_{-1}=K_L$, $S_{N}=K_R$, $O_{-1}=Y^0_P-B_L*K_L$, $O_{N}=Y^N_P-B_R*K_R$.
After this, the overall inference computation is simplified as: calculate $idx^{'}$ then $x*S_{idx^{'}} + O_{idx^{'}}$.
We implement it in CUDA and test the inference time as shown in Table \ref{inference_time}.
As expected, PWLU is very efficient at inference, close to ReLU and Swish.
As for training, PWLU is around 20\% slower compared to ReLU due to more gradient terms.
\begin{table}[h]
\centering
\small
\begin{tabular}{p{1.5cm}<{\centering}p{0.6cm}<{\centering}p{0.6cm}<{\centering}p{0.7cm}<{\centering}p{0.8cm}<{\centering}p{0.5cm}<{\centering}p{0.5cm}}
\toprule
  Model      & ReLU & Swish & PWLU & FReLU & APL  & PAU \\
\midrule
ResNet-18    & 17.1 & 18.8  & 19.9 & 23.3   & 40.7 & 64.2 \\
MobileNetV3  & 15.4 & 15.7  & 16.2 & 18.1   & 31.7 & 38.6 \\
\bottomrule
\end{tabular}
\vspace{0.2cm}
\caption{Inference time(ms) measured on an NVIDIA V100 GPU with input size
of $64\times3\times224\times224$, averaged for 500 times.}
\label{inference_time}
\end{table}

\vspace{-0.4cm}

\subsection{Ablation on input-boundary misalignment}

We further study the influences of the input-boundary misalignment problem introduced in Section \ref{sec_search}.
According to our analysis, the $\left[B_L, B_R\right]$ and  input distribution of PWLU can be misaligned during normal training, wasting the flexibility of PWLU and affecting the final performance.
The statistic-based realignment is designed to address this problem,so we compare these two methods:
\begin{itemize}
\item \emph{fix-init-$x$}: intialize $\left[B_L, B_R\right]$ to a hand-picked $\left[-x, x\right]$ then do normal training
\item \emph{stat-realign}: train with statistic-based realignment according to Figure \ref{two-stage}
\end{itemize}
We run experiments on ResNet-18 and MobileNet-V3 for both layer-wise and channel-wise PWLU. We choose $N=16$ for PWLU in all experiments.

\begin{table}[h]
\centering
\small
\begin{tabular}{lcc}
\toprule
  Method         & layer-wise & channel-wise  \\
\midrule
ResNet-18 baseline  & \multicolumn{2}{c}{71.34}   \\
+ PWLU-fix-init-3  & 71.57     & 72.27 \\
+ PWLU-fix-init-6  & 71.70     & 71.86 \\
+ PWLU-fix-init-10 & 71.84     & 72.19 \\
+ PWLU-stat-realign   & \textbf{72.22}     & \textbf{72.57} \\
\bottomrule
\end{tabular}
\vspace{0.2cm}
\caption{Top-1 accuracies on ImageNet for ResNet-18}
\label{tab-bd-init-res18}
\end{table}

\begin{table}[h]
\centering
\small
\begin{tabular}{lcc}
\toprule
  Method         & layer-wise & channel-wise \\
\midrule
MobileNet-V3 baseline  & \multicolumn{2}{c}{67.74} \\
+ PWLU-fix-init-3  & 68.54    & 68.98  \\
+ PWLU-fix-init-6  & 68.48    & 68.88  \\
+ PWLU-fix-init-10 & 68.42    & 68.20  \\
+ PWLU-stat-realign   & \textbf{69.30}    & \textbf{69.65} \\
\bottomrule
\end{tabular}
\vspace{0.2cm}
\caption{Top-1 accuracies on ImageNet for MobileNet-V3}
\label{tab-bd-init-mbv3}
\end{table}

As shown in Table \ref{tab-bd-init-res18} and Table \ref{tab-bd-init-mbv3}, layer-wise PWLU with \emph{fix-init} shows slightly improvements over baselines.
When using \emph{stat-realign}, the performance further boosts by a clear margin.
Note that channel-wise PWLU is more sensitive to different initial values of \emph{fix-init}, while \emph{stat-realign} consistently achieves better results.

\begin{figure}[h]
\includegraphics[width=0.45\textwidth]{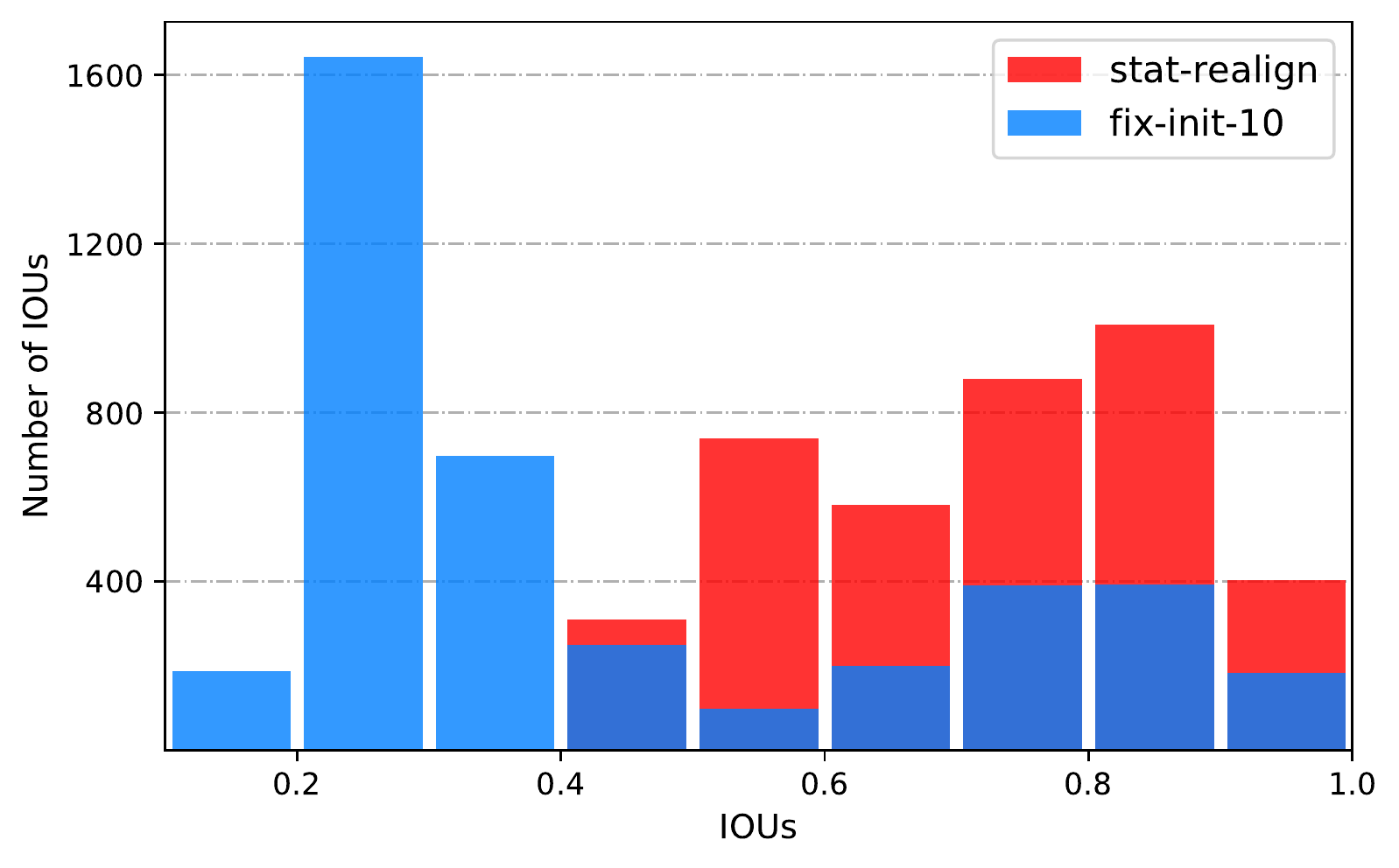}
\caption{Distribution of IOUs computed for PWLU-fix-init-10 and PWLU-stat-realign}
\label{bd-fix-vs-stat-ch}
\end{figure}

To further verify our analysis, we quantitatively show the alignment between $\left[B_L, B_R\right]$ and the input distribution of PWLU.
At the end of training, we collect input $x$ of each single channel of all PWLUs, then compute IOU for $\left[B_L, B_R\right]$ and $\left[x_{\text{p}0.05}, x_{\text{p}0.95}\right]$:
\begin{equation}
\text{IOU} = \frac{\left[B_L, B_R\right] \cap \left[x_{\text{p}0.05}, x_{\text{p}0.95}\right]}{\left[B_L, B_R\right] \cup \left[x_{\text{p}0.05}, x_{\text{p}0.95}\right]}
\end{equation}
This IOU can be regarded as a quantitative measure of alignment at each channel.
We choose two experiments in Table \ref{tab-bd-init-mbv3}: PWLU-fix-init-10 and PWLU-stat-realign under channel-wise setting, compute IOUs for each experiment and show their distributions in Figure \ref{bd-fix-vs-stat-ch}.
It clearly shows that \emph{stat-realign} has much higher IOUs compared to \emph{fix-init}, successfully addressing the input-boundary misalignment problem.
Based on the above results, we can conclude that the input-boundary misalignment problem indeed affects the learning of PWLU, and it can be addressed by our statistic-based realignment method.

\subsection{Ablation on interval numbers}
As described in Section \ref{definition}, the number of intervals $N$ is a hyperparameter.
A larger number of intervals brings higher degrees of freedom, increasing the capacity of models.
Here we show the influence of $N$, to facilitate the choice of this hyperparameter.

We run experiments for $N \in \{4, 8, 12, 16, 20\}$, for three architectures with channel-wise PWLU.
As shown in Table \ref{tab-n}, more intervals generally bring greater improvements especially for small models.
For example, PWLU with $N=8$ shows 0.8\% improvements over $N=4$ for MobileNet-V3.
One interesting observation is that for all three models, $N=20$ shows slightly worse results.
A possible explanation is that each $Y_P^{idx}$ only influences the computation of adjacent two intervals.
When $N$ grows larger, each interval gets a fewer number of input data points, which will reduce the number of corresponding gradients received by $Y_P^{idx}$.
This may result in higher variances of aggregated gradients and influence the training.
Generally, $N \in \left[8, 16\right]$ gives good results, which is the recommended range for $N$.

\begin{table}[h]
\centering
\small
\begin{tabular}{cccc}
\toprule
  $N$  & ResNet-18 & MobileNet-V3 & EfficientNet-B0 \\
\midrule
4      & 72.24     & 68.70        & 77.90 \\
8      & 72.39     & 69.55        & 78.09 \\
12     & 72.45     & 69.20        & 77.96 \\
16     & \textbf{72.57}     & \textbf{69.65}        & \textbf{78.22}  \\
20     & 72.43     & 69.37        & 77.86  \\
\bottomrule
\end{tabular}
\vspace{0.2cm}
\caption{Top-1 accuracies on ImageNet for different number of intervals.}
\label{tab-n}
\end{table}

\subsection{Comparing to APL and PAU}

Both APL and PAU use parameterized formulations covering a wide range of scalar functions, similar to PWLU.
However, they may have large differences in optimization.
For example, APL is a weighted sum of many basis functions: \text{APL}(x) = $\text{max}(0, x) + \sum_{i=1}^S a_i\text{max}(0, -x+b_i)$.
One can clearly see that $a_i$ and $b_i$ only receive gradients when $x<b_i$.
Thus the gradients with respect to $a_i$ and $b_i$ can be highly unbalanced depending on the value of $b_i$.
A larger value of $b_i$ enables more gradients and vice versa.
This unbalance makes the optimization of APL difficult, and we indeed find it unstable in our experiments.
PAU uses a more sophisticated formulation based on Pad\'e approximant.
It works well sometimes but improvements are not as consistent as PWLU.
One possible explanation is that PAU defines a global function for $\mathbb{R}$ but the actual input distribution only lies in a bounded region.
This gap may reduce the expressiveness of PAU and weakens the improvements.
Comparing to APL and PAU, our PWLU has balanced gradients, focuses on bounded regions and carefully handles the input-boundary misalignment problem.
These designs make PWLU consistently outperforms APL and PAU across different architectures.

\subsection{Visualization for learned PWLUs}

\vspace{-0.3cm}

\begin{figure}[h]
\centering
\includegraphics[width=0.48\textwidth]{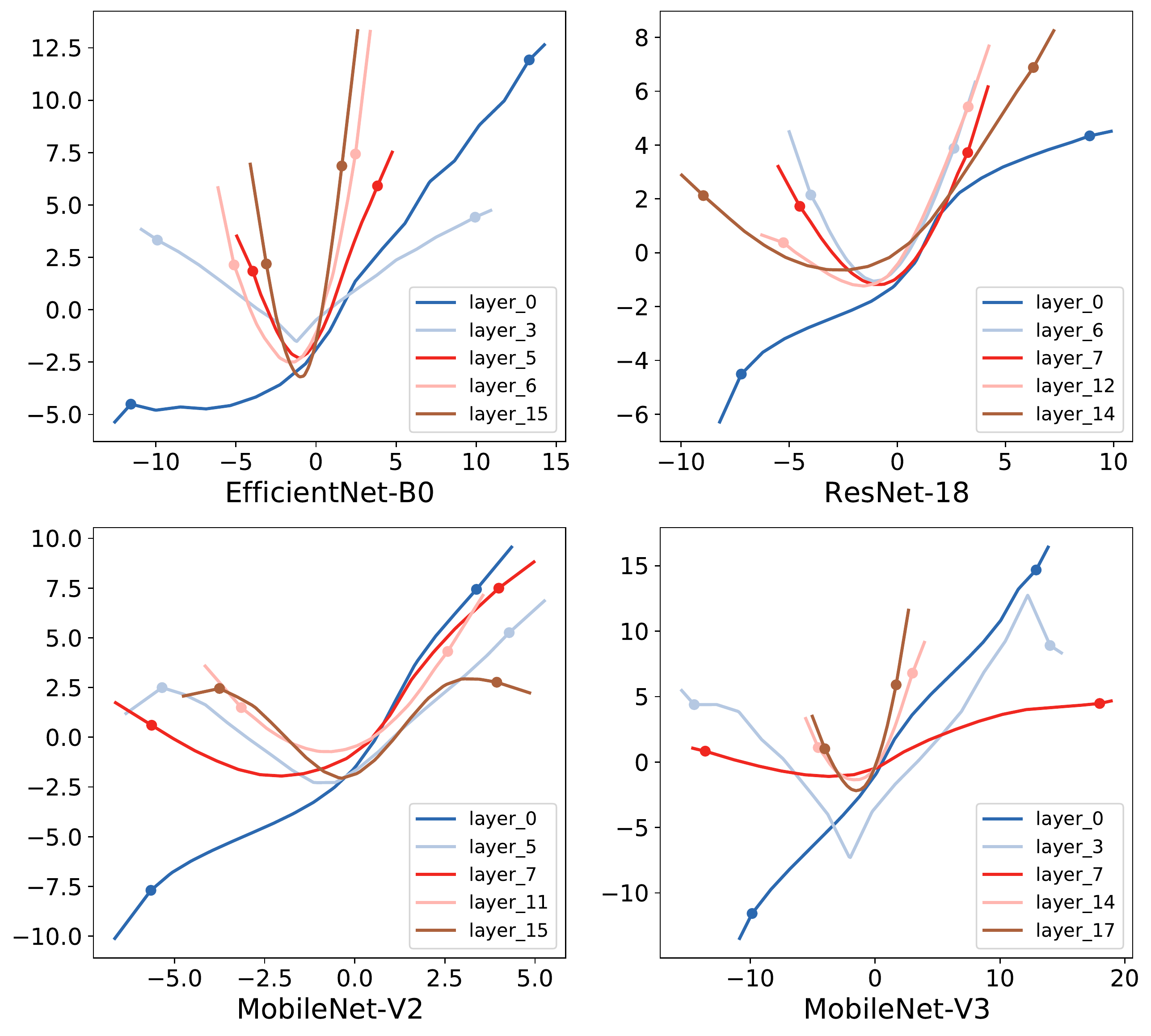}
\caption{Learned PWLUs for different layers and architectures. Only five functions are sampled for clarity. Boundaries are shown by markers with each function.}
\label{visualize}
\end{figure}

We visualize learned PWLUs for different layers and architectures in Figure \ref{visualize}.
It clearly shows that learned activation functions are substantially different from hand-designed ones.
V-shaped functions frequently appear, which are rarely seen in common activation functions.
Also, activations for the first layer are close to linear functions.
While sharing some common patterns, the learned functions are clearly different across architectures,
which is consistent with our motivation of learning specialized activation functions for different architectures.

\section{Conclusion}

In this paper, we propose a new activation function called PWLU, which is designed to learn specialized activation functions for each dataset and architecture.
To achieve this goal, we carefully design the formulation and the learning method.
The piecewise-linear-based formulation is flexible and can be easily learned by gradients, while the statistic-based realignment makes the learning more effective and robust.
On large-scale datasets like ImageNet and COCO, PWLU consistently outperforms other activation functions across multiple architectures, which demonstrates the benefits of learning specialized activation functions.
PWLU is both simple in implementation and efficient at inference, which has great potential in real-world applications.

{\small
\bibliographystyle{ieee_fullname}
\bibliography{egbib}

\begin{thebibliography}{10}\itemsep=-1pt

\bibitem{agostinelli2014learning}
Forest Agostinelli, Matthew Hoffman, Peter Sadowski, and Pierre Baldi.
\newblock Learning activation functions to improve deep neural networks.
\newblock {\em arXiv preprint arXiv:1412.6830}, 2014.

\bibitem{bello2017neural}
Irwan Bello, Barret Zoph, Vijay Vasudevan, and Quoc~V Le.
\newblock Neural optimizer search with reinforcement learning.
\newblock {\em arXiv preprint arXiv:1709.07417}, 2017.

\bibitem{chen2020dynamic}
Yinpeng Chen, Xiyang Dai, Mengchen Liu, Dongdong Chen, Lu Yuan, and Zicheng
  Liu.
\newblock Dynamic relu.
\newblock {\em arXiv preprint arXiv:2003.10027}, 2020.

\bibitem{clevert2015fast}
Djork-Arn{\'e} Clevert, Thomas Unterthiner, and Sepp Hochreiter.
\newblock Fast and accurate deep network learning by exponential linear units
  (elus).
\newblock {\em arXiv preprint arXiv:1511.07289}, 2015.

\bibitem{dauphin2017language}
Yann~N Dauphin, Angela Fan, Michael Auli, and David Grangier.
\newblock Language modeling with gated convolutional networks.
\newblock In {\em International conference on machine learning}, pages
  933--941, 2017.

\bibitem{goodfellow2013maxout}
Ian Goodfellow, David Warde-Farley, Mehdi Mirza, Aaron Courville, and Yoshua
  Bengio.
\newblock Maxout networks.
\newblock In {\em International conference on machine learning}, pages
  1319--1327. PMLR, 2013.

\bibitem{hahnloser2000digital}
Richard~HR Hahnloser, Rahul Sarpeshkar, Misha~A Mahowald, Rodney~J Douglas, and
  H~Sebastian Seung.
\newblock Digital selection and analogue amplification coexist in a
  cortex-inspired silicon circuit.
\newblock {\em Nature}, 405(6789):947--951, 2000.

\bibitem{he2019rethinking}
Kaiming He, Ross Girshick, and Piotr Doll{\'a}r.
\newblock Rethinking imagenet pre-training.
\newblock In {\em Proceedings of the IEEE international conference on computer
  vision}, pages 4918--4927, 2019.

\bibitem{he2017mask}
Kaiming He, Georgia Gkioxari, Piotr Doll{\'a}r, and Ross Girshick.
\newblock Mask r-cnn.
\newblock In {\em Proceedings of the IEEE international conference on computer
  vision}, pages 2961--2969, 2017.

\bibitem{he2015delving}
Kaiming He, Xiangyu Zhang, Shaoqing Ren, and Jian Sun.
\newblock Delving deep into rectifiers: Surpassing human-level performance on
  imagenet classification.
\newblock In {\em Proceedings of the IEEE international conference on computer
  vision}, pages 1026--1034, 2015.

\bibitem{he2016deep}
Kaiming He, Xiangyu Zhang, Shaoqing Ren, and Jian Sun.
\newblock Deep residual learning for image recognition.
\newblock In {\em Proceedings of the IEEE conference on computer vision and
  pattern recognition}, pages 770--778, 2016.

\bibitem{hochreiter1997long}
Sepp Hochreiter and J{\"u}rgen Schmidhuber.
\newblock Long short-term memory.
\newblock {\em Neural computation}, 9(8):1735--1780, 1997.

\bibitem{howard2019searching}
Andrew Howard, Mark Sandler, Grace Chu, Liang-Chieh Chen, Bo Chen, Mingxing
  Tan, Weijun Wang, Yukun Zhu, Ruoming Pang, Vijay Vasudevan, et~al.
\newblock Searching for mobilenetv3.
\newblock In {\em Proceedings of the IEEE International Conference on Computer
  Vision}, pages 1314--1324, 2019.

\bibitem{hu2018squeeze}
Jie Hu, Li Shen, and Gang Sun.
\newblock Squeeze-and-excitation networks.
\newblock In {\em Proceedings of the IEEE conference on computer vision and
  pattern recognition}, pages 7132--7141, 2018.

\bibitem{jarrett2009best}
Kevin Jarrett, Koray Kavukcuoglu, Marc'Aurelio Ranzato, and Yann LeCun.
\newblock What is the best multi-stage architecture for object recognition?
\newblock In {\em 2009 IEEE 12th international conference on computer vision},
  pages 2146--2153. IEEE, 2009.

\bibitem{klambauer2017self}
G{\"u}nter Klambauer, Thomas Unterthiner, Andreas Mayr, and Sepp Hochreiter.
\newblock Self-normalizing neural networks.
\newblock In {\em Advances in neural information processing systems}, pages
  971--980, 2017.

\bibitem{krizhevsky2009learning}
Alex Krizhevsky, Geoffrey Hinton, et~al.
\newblock Learning multiple layers of features from tiny images.
\newblock 2009.

\bibitem{alexnet}
Alex Krizhevsky, Ilya Sutskever, and Geoffrey~E Hinton.
\newblock Imagenet classification with deep convolutional neural networks.
\newblock In F. Pereira, C.~J.~C. Burges, L. Bottou, and K.~Q. Weinberger,
  editors, {\em Advances in Neural Information Processing Systems}, volume~25,
  pages 1097--1105. Curran Associates, Inc., 2012.

\bibitem{lin2017feature}
Tsung-Yi Lin, Piotr Doll{\'a}r, Ross Girshick, Kaiming He, Bharath Hariharan,
  and Serge Belongie.
\newblock Feature pyramid networks for object detection.
\newblock In {\em Proceedings of the IEEE conference on computer vision and
  pattern recognition}, pages 2117--2125, 2017.

\bibitem{lin2017focal}
Tsung-Yi Lin, Priya Goyal, Ross Girshick, Kaiming He, and Piotr Doll{\'a}r.
\newblock Focal loss for dense object detection.
\newblock In {\em Proceedings of the IEEE international conference on computer
  vision}, pages 2980--2988, 2017.

\bibitem{lin2014microsoft}
Tsung-Yi Lin, Michael Maire, Serge Belongie, James Hays, Pietro Perona, Deva
  Ramanan, Piotr Doll{\'a}r, and C~Lawrence Zitnick.
\newblock Microsoft coco: Common objects in context.
\newblock In {\em European conference on computer vision}, pages 740--755.
  Springer, 2014.

\bibitem{ma2020funnel}
Ningning Ma, Xiangyu Zhang, and Jian Sun.
\newblock Funnel activation for visual recognition.
\newblock {\em arXiv preprint arXiv:2007.11824}, 2020.

\bibitem{maas2013rectifier}
Andrew~L Maas, Awni~Y Hannun, and Andrew~Y Ng.
\newblock Rectifier nonlinearities improve neural network acoustic models.
\newblock In {\em Proc. icml}, volume~30, page~3, 2013.

\bibitem{molina2020pade}
Alejandro Molina, Patrick Schramowski, and Kristian Kersting.
\newblock Pad\'e activation units: End-to-end learning of flexible activation
  functions in deep networks, 2020.

\bibitem{nair2010rectified}
Vinod Nair and Geoffrey~E Hinton.
\newblock Rectified linear units improve restricted boltzmann machines.
\newblock In {\em ICML}, 2010.

\bibitem{ramachandran2017searching}
Prajit Ramachandran, Barret Zoph, and Quoc~V Le.
\newblock Searching for activation functions.
\newblock {\em arXiv preprint arXiv:1710.05941}, 2017.

\bibitem{ren2015faster}
Shaoqing Ren, Kaiming He, Ross Girshick, and Jian Sun.
\newblock Faster r-cnn: Towards real-time object detection with region proposal
  networks.
\newblock In {\em Advances in neural information processing systems}, pages
  91--99, 2015.

\bibitem{russakovsky2015imagenet}
Olga Russakovsky, Jia Deng, Hao Su, Jonathan Krause, Sanjeev Satheesh, Sean Ma,
  Zhiheng Huang, Andrej Karpathy, Aditya Khosla, Michael Bernstein, et~al.
\newblock Imagenet large scale visual recognition challenge.
\newblock {\em International journal of computer vision}, 115(3):211--252,
  2015.

\bibitem{sandler2018mobilenetv2}
Mark Sandler, Andrew Howard, Menglong Zhu, Andrey Zhmoginov, and Liang-Chieh
  Chen.
\newblock Mobilenetv2: Inverted residuals and linear bottlenecks.
\newblock In {\em Proceedings of the IEEE conference on computer vision and
  pattern recognition}, pages 4510--4520, 2018.

\bibitem{srivastava2015highway}
Rupesh~Kumar Srivastava, Klaus Greff, and J{\"u}rgen Schmidhuber.
\newblock Highway networks.
\newblock {\em arXiv preprint arXiv:1505.00387}, 2015.

\bibitem{tan2019efficientnet}
Mingxing Tan and Quoc~V Le.
\newblock Efficientnet: Rethinking model scaling for convolutional neural
  networks.
\newblock {\em arXiv preprint arXiv:1905.11946}, 2019.

\bibitem{van2016conditional}
Aaron Van~den Oord, Nal Kalchbrenner, Lasse Espeholt, Oriol Vinyals, Alex
  Graves, et~al.
\newblock Conditional image generation with pixelcnn decoders.
\newblock In {\em Advances in neural information processing systems}, pages
  4790--4798, 2016.

\end{thebibliography}
}

\end{document}